%% file: main.tex
\pdfoutput=1
\documentclass[sigconf]{acmart}
\usepackage{url}
\usepackage{amsmath}
\usepackage{algorithm}
\usepackage{algorithmicx}
\usepackage{algpseudocode}
\usepackage{mathrsfs}
\usepackage{amsmath}
\usepackage{subfigure}
\usepackage{bm}
\usepackage{caption}
\usepackage{url}            
\usepackage{booktabs}       
\usepackage{amsfonts}       
\usepackage{nicefrac}       
\usepackage{microtype}      
\usepackage{xcolor}         
\usepackage{subfigure}
\usepackage{graphicx}
\usepackage{bm}
\usepackage{amsmath}
\usepackage{wrapfig}
\usepackage{threeparttable}
\usepackage{amsthm}

\usepackage{algorithm}
\usepackage{algpseudocode}
\usepackage{multirow}
\usepackage{makecell}
\usepackage{balance}
\usepackage{wrapfig}
\usepackage{float}
\usepackage{xspace}
\usepackage{enumitem}

\input{math_commands.tex}

\AtBeginDocument{%
  \providecommand\BibTeX{{%
    Bib\TeX}}}

\copyrightyear{2025}
\acmYear{2025}
\setcopyright{acmlicensed}
\acmConference[KDD '25] {Proceedings of the 31st ACM SIGKDD Conference on Knowledge Discovery and Data Mining V.1}{August 3--7, 2025}{Toronto, ON, Canada.}
\acmBooktitle{Proceedings of the 31st ACM SIGKDD Conference on Knowledge Discovery and Data Mining V.1 (KDD '25), August 3--7, 2025, Toronto, ON, Canada}
\acmISBN{979-8-4007-1245-6/25/08}
\acmDOI{10.1145/3690624.3709260}

\begin{document}

\title[IN-Flow: Instance Normalization Flow for Non-stationary Time Series Forecasting]{IN-Flow: Instance Normalization Flow for Non-stationary \\ Time Series Forecasting}

\author{Wei Fan}
\email{weifan.oxford@gmail.com}
\affiliation{%
  \institution{University of Oxford}
  \city{Oxford}
  \country{UK}
}

\author{Shun Zheng}
\email{shun.zheng@microsoft.com}
\affiliation{%
  \institution{Microsoft Research Asia}
  \city{Beijing}
  \country{China}
}

\author{Pengyang Wang}
\authornote{Corresponding Author.}
\email{pywang@um.edu.mo}
\affiliation{%
  \institution{University of Macau}
  \city{Macau SAR}
  \country{China}
}

\author{Rui Xie}
\email{rui.xie@ucf.edu}
\affiliation{%
  \institution{University of Central Florida}
  \city{Orlando}
  \country{US}
}

\author{Kun Yi}
\email{kunyi.cn@gmail.com}
\affiliation{%
  \institution{State Information Center}
  \city{Beijing}
  \country{China}
}

\author{Qi Zhang}
\email{zhangqi_cs@tongji.edu.cn}
\affiliation{%
  \institution{Tongji University}
  \city{Shanghai}
  \country{China}
}

\author{Jiang Bian}
\email{jiang.bian@microsoft.com}
\affiliation{%
  \institution{Microsoft Research Asia}
  \city{Beijing}
  \country{China}
}

\author{Yanjie Fu}
\email{yanjie.fu@asu.edu}
\affiliation{%
  \institution{Arizona State University}
  \city{Tempe}
  \country{US}
}

\renewcommand{\shortauthors}{Wei Fan et al.}

\begin{abstract}
  \input{0_abstract}
\end{abstract}





\maketitle

\input{1_introduction}
\input{2_related_work}

\input{3_preliminary}
\input{4_method}
\input{5_experiments}
\input{6_conclusion}

\section{Acknowledgments}
Most of this work is done when the first author was a research assistant in State Key Laboratory of Internet of Things for Smart City at the University of Macau.
This research is funded by the Science and Technology Development Fund (FDCT), Macau SAR (file no. 0123/2023/RIA2, 001/2024/SKL).

\bibliographystyle{ACM-Reference-Format}
\bibliography{ref}

\input{Appendix}

\end{document}

%% file: math_commands.tex
\usepackage{amsmath,amsfonts,bm}

\def\eqref#1{equation~\ref{#1}}

\def\1{\bm{1}}

\DeclareMathAlphabet{\mathsfit}{\encodingdefault}{\sfdefault}{m}{sl}
\SetMathAlphabet{\mathsfit}{bold}{\encodingdefault}{\sfdefault}{bx}{n}



%% file: 0_abstract.tex
Due to the non-stationarity of time series, the distribution shift problem largely hinders the performance of time series forecasting. Existing solutions either rely on using certain statistics to specify the shift, or developing specific mechanisms for certain network architectures. However, the former would fail for the unknown shift beyond simple statistics, while the latter has limited compatibility on different forecasting models. To overcome these problems, we first propose a \textit{decoupled formulation} for time series forecasting, with no reliance on fixed statistics and no restriction on forecasting architectures. This formulation regards the removing-shift procedure as a special transformation between a raw distribution and a desired target distribution and separates it from the forecasting. Such a formulation is further formalized into a \emph{bi-level optimization} problem, to enable the joint learning of the transformation (outer loop) and forecasting (inner loop). Moreover, the special requirements of expressiveness and bi-direction for the transformation motivate us to propose \textit{instance normalization flow} (IN-Flow), a novel invertible network for time series transformation. Different from the classic ``normalizing flow'' models, IN-Flow does not aim for normalizing input to the prior distribution (e.g., Gaussian distribution) for generation, but creatively transforms time series distribution by stacking normalization layers and flow-based invertible networks, which is thus named ``normalization'' flow. Finally, we have conducted extensive experiments on both synthetic data and real-world data, which demonstrate the superiority of our method. 



%% file: 1_introduction.tex
\vspace{-1mm}
\section{Introduction} \label{sec:intro}
With wide applications in many real-world scenarios, such as electricity consumption planning~\cite{akay2007grey,fan2024dewp}, traffic flow analysis~\cite{lv2014traffic}, and weather condition estimation \cite{abhishek2012weather,han2021joint}, time series forecasting has received broad research attention for decades, with rapidly-evolving learning paradigm from traditional statistical approaches~\cite{holt1957forecasting,holt2004forecasting,winters1960forecasting,whittle1951hypothesis,whittle1963prediction} to modern deep learning based methods~\cite{salinas2020deepar,rangapuram2018deep,Oreshkin2020N-BEATS,zhou2021informer,wu2021autoformer,deng2024parsimony,deng2024disentangling,yi2024fouriergnn,kunyi_2023survey}, resulting in progressively yet significantly improved time series forecasting performance.

It is noteworthy that most of these methods, especially deep learning-based forecasting models, inherently follow a \emph{stationarity} assumption: the temporal dependencies between historical observations and future predictions are stationary such that they can be effectively captured and generalized to future time points.
However, this \emph{stationarity} assumption overlooks the \emph{non-stationarity} of real-world time series data, which can be characterized by continuously shifted joint distribution over time~\cite{kim2022RevIN}. As well-known as the distribution shift problem, it can substantially hinder the performance of modern deep forecasting models~\cite{kim2022RevIN,liu2022NonstatTranformers,fan2023dish} due to the resulting discrepancies of underlying distributions between the training and test data.
Since time series data are usually collected at a high frequency over a long duration, such non-stationary sequences with millions of timesteps might even lead to a worse situation for forecasting~\cite{woo2022deeptime,fan2024deep}.


Nevertheless, reviewing all the achievements in non-stationary time series forecasting, existing studies always explicitly considering the distribution shift issue in time series forecasting either relied on certain statistics, such as rescaling time series with global minimum and maximum \cite{ogasawara2010adaptive} and specifying non-stationary information as the mean, standard deviation, and learnable/sliced statistics~\cite{kim2022RevIN,fan2023dish,liu2023adaptive}, or focused on developing specific mechanisms for certain network architectures to relieve the shift~\cite{du2021ada,liu2022NonstatTranformers}.
The former methods conduct normalizations towards data as pre-processing or in-processing steps, which would probably fail for unknown distribution shifts beyond a simple change of statistics. In contrast, the latter architecture-specific methods have limited compatibility and could be unadaptable for advanced forecasting architectures in the future. 

To provide a common understanding of distribution shift and non-strationary forecasting and effectively address its interfering effects on forecasting models, in this paper, we develop a systematic approach following two crucial principles:
(1) making no assumption that distribution shifts can be quantified by certain statistics; (2) making the method agnostic to the architectures of forecasting models.
To achieve these ends, we develop a {\textit{{decoupled} formulation}} for time series forecasting:
regarding the procedure of removing distribution shift as a special distribution transformation, we separate the functionalities of removing shift and performing forecasting into a \textit{transformation module} and a \textit{forecasting module} respectively, where
the first module is responsible for the transformation between the raw data distribution and the desired target distribution without shifts, and the second module holds the potential to obtain more accurate forecasting on "cleaned" data samples.
Then, we naturally formalize such a decoupled formulation into a \textit{{bi-level} optimization} problem \cite{colson2007overview,franceschi2018bilevel}: in order to enable the joint learning of the two modules, 
we iteratively improve the forecasting ability by minimizing an inner objective (training errors) and optimize the transformation ability by minimizing an outer object (validation errors) that corresponds to the generalization on the shifted data.
This optimization procedure stimulates the transformation module to reduce shifts in a data-driven manner, leading to more stable and stationary time series forecasting.

Though the formulation easily accommodates any forecasting architectures, another critical challenge lies in the design of the transformation module. Some special requirements motivate us to introduce certain inductive biases. 
First, this module should transform distributions effectively (expressiveness) to relieve the shift. Second, the module is required to transform to and fro (bi-direction) between the raw distribution and the target distribution for the input/output of forecasting models.
Such two considerations inspire us to leverage the key idea of \emph{normalizing flows}~\cite{dinh2014nice,dinh2016density,kingma2018glow}, which can reversibly transform data distributions as one kind of \emph{invertible neural networks} \cite{ardizzone2018analyzing}.
To this end, we propose \textit{instance normalization flow} (IN-Flow), a novel invertible network for time series distribution transformation. 
In the domain of time series forecasting, the format of
multi-variate sequences motivates us to integrate {instance normalization}~\cite{ulyanov2016instance,huang2017arbitrary,kim2022RevIN} together with coupling layers~\cite{dinh2014nice,dinh2016density,kingma2018glow}. Different from traditional normalizing flows targeting for the transformation to a prior simple distribution, IN-Flow targets for the transformation to a desired distribution with shifted information largely removed but forecasting-related information preserved, thus facilitating the final forecasting.
In summary, our contributions can be summarized as follows:
\begin{itemize}[leftmargin=*]
    \item We present a \textit{decoupled formulation} for time series forecasting that separates the functionalities of removing shifts and performing forecasting into a distribution transformation module and a forecasting module, which doesn't rely on fixed statistics, and has no restriction on forecasting architectures.
    \item We formalize such a formulation into a \textit{bi-level optimization} problem to enable the joint learning of forecasting and the transformation in a data-driven manner.
    \item We propose a novel invertible neural network, IN-Flow, to fulfill the transformation for a desired distribution with shifted information removed but forecasting-related preserved.
    \item We conduct extensive experiments on both synthetic data and on real-world data, which demonstrate the consistent superiority of IN-Flow with regard to existing methods in non-stationary time series forecasting.
\end{itemize}

%% file: 2_related_work.tex
\vspace{-1mm}
\section{Related Work}
\subsection{{Deep Learning for Time Series Forecasting}}
Time series forecasting has been a longstanding research topic. At an early stage, researchers have proposed statistical modeling approaches, including exponential smoothing \cite{holt1957forecasting} and auto-regressive moving averages (ARMA) \cite{whittle1963prediction}. With the great successes of deep learning, many deep learning models have been developed for time series forecasting. 
In recent years, deep learning-based methods have gained prominence in time series forecasting due to their ability to capture nonlinear and complex correlations~\cite{Lim_2021}. These methods have employed various network architectures to learn temporal dependencies, such as recurrent neural network~\cite{lstnet_2018,salinas2020deepar}, temporal convolution networks~\cite{bai2018,liu2022scinet}, etc.
One branch of methods has applied pure fully-connected neural networks for forecasting. For example, N-BEATS and DEPTS~\cite{Oreshkin2020N-BEATS, fan2022depts} build residual layers on stacked connected layers for forecasting. N-HiTS~\cite{nhits_2022} integrates multi-rate input sampling and hierarchical interpolation with MLPs to enhance univariate forecasting. DLinear~\cite{dlinear_2023} introduces a simple approach using a single-layer linear model to capture temporal relationships between input and output time series data. 
FreTS~\cite{frets_23} utilizes frequency-domain MLP layers for time series forecasting; Filternet~\cite{yi2024filternet} adopts the frequency filters with the linear layers for the forecasting.
Another branch of methods are built upon Transformer \cite{vaswani2017attention} for the time series forecasting task. 
Many research have improved vanilla transformer
in attention computation, memory consumption, etc, to enhance forecasting \cite{zhou2021informer,wu2021autoformer,zhou2022fedformer}.
To capture intricate dependencies and long-range interactions. 
PatchTST~\cite{patchtst_2023} segments time series into patches as input tokens to the Transformer and maintaining channel independence.
iTransformer~\cite{itransformer_2024} inverts the Transformer's structure by treating independent series as variate tokens to capture multivariate correlations. 

\section{Preliminary}
\label{sec:background}

\vspace{-1mm}
\subsection{{Non-stationary Time Series Forecasting}}
Time series forecasting suffers from the non-stationarity and the distribution shift issue considering distributions of real-world series change over time \cite{akay2007grey}. 
To overcome this problem, some specific mechanisms have been designed for certain network architectures to relieve the distribution shift: for example, \cite{du2021ada} proposed Adaptive RNNs to handle the shift problem on top of recurrent neural networks (RNNs); Nonstationary Transformers \cite{liu2022NonstatTranformers} have proposed series stationarization and de-stationary attention to enhance transformer-based forecasting models. These methods, however, are network-specific architectures that have limited compatibility with future advanced forecasting methods.
Another kind of method for the non-stationarity problem is to utilize the normalization techniques. Usually, normalization is model-agnostic and can be coupled with different backbones. For example, the pioneer work, Adaptive Norm \cite{ogasawara2010adaptive} rescale time series through global minimum and maximum. DAIN \cite{passalis2019deep} use adaptive z-score normalization by global average and std. for the task of time series classification. 
Recently, RevIN \cite{kim2022RevIN} proposes to use instance normalization to reduce time series distribution shift for deep learning time series forecasting methods. Dish-TS~\cite{fan2023dish} considers both the inter-space shift and intra-space shift and proposes coefficient networks to capture learnable statistics for normalization. SAN~\cite{liu2023adaptive} considers the sliced statistics of time series and proposes the corresponding adaptive sliced normalization.
However, these normalization techniques rely on statistics calculation, which fall short in overcoming unknown shifts beyond the statistics.

%% file: 3_preliminary.tex
\subsection{Time Series Forecasting}
\label{sec:bg_tsf}
Let $\bm{s} = [ \bm{s}_1; \bm{s}_2; \cdots; \bm{s}_T ] \in \mathbb{R}^{T \times D}$ be regularly sampled multi-variate time series with $T$ timestamps and $D$ variates, where $\bm{s}_t \in \mathbb{R}^D$ denotes the multi-variate values at timestamp $t$.
Besides, we use $\bm{x}^L_t \in \mathbb{R}^{L \times D}$ to denote a length-$L$ segment of $\bm{s}$ ending at timestamp $t$ (exclusive), namely $\bm{x}^L_t = \bm{s}_{t-L:t} = [\bm{s}_{t-L}; \bm{s}_{t-L+1}; \cdots; \bm{s}_{t-1}]$.
Similarly, we represent a length-$H$ segment of $\bm{s}$ starting from timestamp $t$ (inclusive) as $\bm{y}^H_t$, so we have $\bm{y}^H_t = \bm{s}_{t:t+H} = [\bm{s}_{t}; \bm{s}_{t+1}; \cdots; \bm{s}_{t+H-1}]$.
The classic time series forecasting formulation is to project
historical observations $\bm{x}^L_t$ into their subsequent future values ${\bm{y}}^H_t$.
Specifically, a typical forecasting model $f_\theta : \mathbb{R}^{L \times D} \to \mathbb{R}^{H \times D}$ produces forecasts by $\hat{\bm{y}}^H_t = f_\theta( \bm{x}^H_t )$ where $\hat{\bm{y}}^H_t$ stands for the forecasting result, $\theta$ encapsulates the model parameters, and $H$ and $L$ are usually referred to as the lengths of horizon (lookahead) windows and lookback windows, respectively.

\vspace{-1mm}
\subsection{Bi-level Optimization}
\label{sec:bg_bilevel_optim}
Bi-level optimization problems are one kind of optimization problems in which a set of variables in an objective function are constrained by the optimal solution of another optimization problem \cite{colson2007overview}. Specifically, given two functions $J_{in}$ and $J_{out}$ as the inner objectives and outer objectives, we consider two set of variables $\theta \in \mathbb{R}^m, \phi \in \mathbb{R}^n$ as the inner variables and outer variables, respectively; the bi-level problem can be given by
$\underset{\phi, \theta_\phi}{\min} J_{out}(\theta_\phi,\phi)$ such that $\theta_\phi \in \arg \underset{\theta}{\min} J_{in}(\theta, \phi) $.
Bi-level problems have involved in many applications, including hyperparmeter optimization, multi-task, and meta-learning \cite{colson2007overview,sinha2017review,franceschi2018bilevel,rajeswaran2019meta}.

\vspace{-1mm}
\subsection{Normalzing Flows} 
\label{sec:bg_flows}
Normalzing flows \cite{papamakarios2021normalizing}, as one common kind of invertible neural networks, can conduct transformation between distributions. Given a high-dimensional random variable $\mathbf{x} \in \mathcal{X}$, flow-based models \cite{dinh2014nice,dinh2016density,papamakarios2017masked,kingma2018glow} transform densities $p_{\mathcal{X}}$ into certain target distribution $p_{\mathcal{Z}}$ (e.g., an isotropic Gaussian) on a latent variable $\mathbf{z} \in \mathcal{Z}$. The transformation is a bijection $g: \mathcal{X} \to \mathcal{Z}$ (with inverse $ g^{-1}$). 
The probability density can be written by the change of variable formula as:
$p_{\mathcal{X}}(\mathbf{x}) = p_{\mathcal{Z}}(\mathbf{z}) \left|\operatorname{det} \left( \frac{\partial g(\mathbf{x})}{\partial \mathbf{x}} \right) \right|$
where $\partial g(\mathbf{x}) / \partial \mathbf{x}$ is the Jacobian of $g$ at $\mathbf{x}$. The transformation $g$ is composed of a sequence of differentiable invertible (or bijective) mapping functions $g = g_1 \circ g_2 \circ \cdots \circ g_P$ that map 
$\mathbf{x} \stackrel{g_1}{\longleftrightarrow} \mathbf{h}_1 \stackrel{g_2}{\longleftrightarrow}  \mathbf{h}_2 \cdots \stackrel{g_P}{\longleftrightarrow} \mathbf{z}$ reversibly.  
Usually, flow-based models are trained via maximum likelihood on training data. Thus, the probability density function of the flows given a data point is
$\log p_{\mathcal{X}}(\mathbf{x}) = \log p_{\mathcal{Z}}(\mathbf{z}) + \log \left|\operatorname{det} \left( \frac{\partial \mathbf{z}}{\partial \mathbf{x}} \right) \right| =\log p_{\mathcal{Z}}(\mathbf{z})+\sum_{i=1}^P  
\log \left| \operatorname{det}\left( \partial \mathbf{h}_i / \partial \mathbf{h}_{i-1} \right) \right|$, where  $\mathbf{h}_0 \triangleq \mathbf{x}$ and $\mathbf{h}_P \triangleq \mathbf{z}$; $\log \left| \operatorname{det}\left( \partial \mathbf{h}_i / \partial \mathbf{h}_{i-1} \right) \right|$ is called log-determinant of $\left( \partial \mathbf{h}_i / \partial \mathbf{h}_{i-1} \right)$. The key idea of designing transformation in normalizing flows is to make the computation of log-determinant simple and easy \cite{kobyzev2020normalizing,papamakarios2021normalizing}.

%% file: 4_method.tex
\begin{figure*}[!t]
\centering
\includegraphics[width=0.96\linewidth]{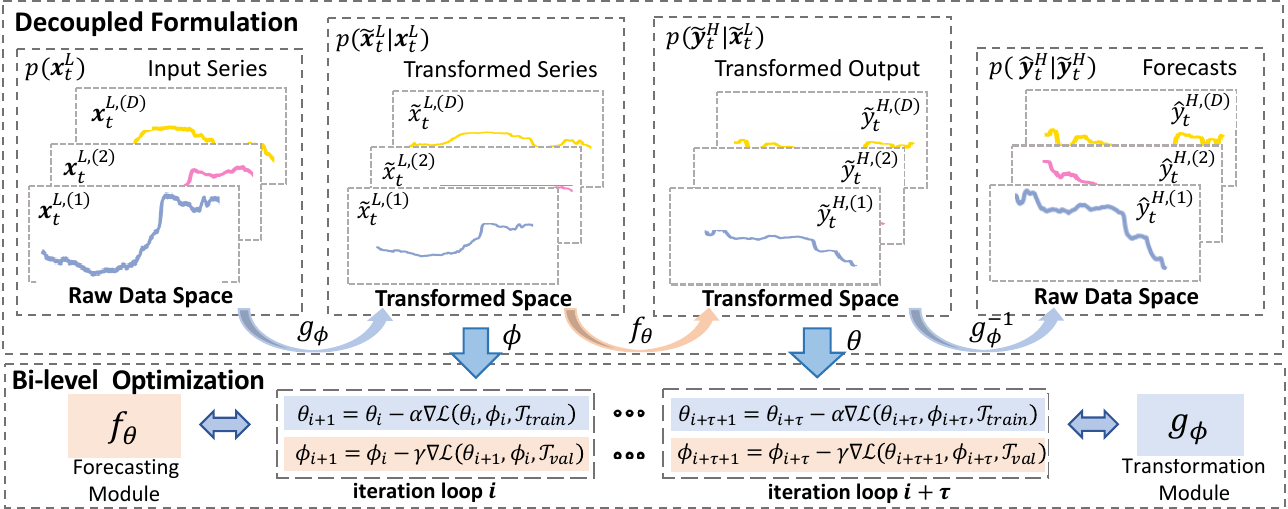}
\vspace{-1mm}
\caption{Framework overview. The upper part is the proposed decoupled formulation (Section \ref{sec:method_formu}): it separates the time series forecasting into the transformation module ($g_\phi$) and the forecasting module ($f_\theta$); the input series with the raw data space is converted into the transformed space by $g_\phi$ for the forecasting conducted by the forecasting module $f_\theta$; then the predicted results will be recovered by the inverse of $g_\phi$ to recovered to the raw data space.
The lower part is the bi-level optimization (Section \ref{sec:method_optim}) for the decoupled formulation of time series forecasting.} \label{fig:framework_overview}
\vspace{-0mm}
\end{figure*}

\section{Methodology}
\label{sec:method}

In this section, we elaborate on our principled approach that makes no assumption to specify distribution shifts and accommodates all kinds of forecasting models.
First, we introduce a decoupled formulation in Section \ref{sec:method_formu}; 
then, we formalize a bi-level optimization problem for the decoupled formulation in Section \ref{sec:method_optim}, and present our specifically designed IN-Flow model in Section~\ref{sec:method_in_flow}.
Besides, Figure \ref{fig:framework_overview} demonstrates an overview of our framework.

\subsection{The Novel Decoupled Formulation}
\label{sec:method_formu}

It is noteworthy that the aforementioned formulation ($\hat{\bm{y}}^H_t = f_{\theta} ( \bm{x}^H_t )$),
embracing a large body of forecasting models~\cite{Oreshkin2020N-BEATS,zhou2021informer,wu2021autoformer,minhao2022SCINet,zeng2022D-linear},
essentially follows a \emph{stationary} assumption:
\emph{the conditional data probability $P(\bm{y}^H_t | \bm{x}^L_t)$ does not change over time}.
However, in practice, some recent studies~\cite{kim2022RevIN,du2021ada,liu2022NonstatTranformers} have well recognized the wide existence of non-stationary properties in real-world time series.
Besides, we further note that those distribution shift patterns can be far more sophisticated than simple statistical measures, such as mean and standard deviation used in~\cite{kim2022RevIN}.
Thus arbitrarily specifying non-stationary properties has the risk of removing certain information crucial to forecasting and is limited in tackling more diverse distribution shifts.

To provide a generic approach to address the problem of distribution shift in time-series forecasting, we regard the procedure of removing non-stationary information as a special distribution transformation (\textit{transformation module}) and separate its functionality from the \textit{forecasting module}.
Accordingly, we develop a decoupled formulation as:
\begin{equation}
    p(\bm{y}^H_t | \bm{x}^L_t) = 
    p(\bm{y}^H_t | \tilde{\bm{y}}^H_t) \cdot 
    p(\tilde{\bm{y}}^H_t | \tilde{\bm{x}}^L_t)  \cdot 
    p(\tilde{\bm{x}}^L_t | \bm{x}^L_t)
    \label{eq:decouple_formula}
\end{equation}
where $\tilde{\bm{x}}^L_t$ and $\tilde{\bm{y}}^H_t$ denote the transformed variables of $\bm{x}^L_t$ and $\bm{y}^H_t$, respectively.
We assume that $\tilde{\bm{x}}^L_t$ and $\tilde{\bm{y}}^H_t$ come from certain desired distributions with shifted information removed but forecasting-related information preserved.
In this way, we decouple the modeling of conditional data distribution $p(\bm{y}^H_t | \bm{x}^L_t)$ into three parts:
(i) $p(\tilde{\bm{x}}^L_t | \bm{x}^L_t)$, performing a transformation on raw input series to remove non-stationary (shifted) information;
(ii) $p(\tilde{\bm{y}}^H_t | \tilde{\bm{x}}^L_t)$, conducting forecasting on the transformed space;
(iii) $p(\bm{y}^H_t | \tilde{\bm{y}}^H_t)$, performing a reverse transformation to recover certain non-stationary patterns and produce final forecasts.

This decoupled formulation lays the foundation for respective parameterization and optimization of transformation and forecasting modules.
Then in addition to the forecasting function $f_\theta$, we introduce another two functions to be responsible for the bi-directional distribution transformations in the transformation module.
To be specific, $g_{\phi^L}: \bm{x}^L_t \rightarrow \tilde{\bm{x}}^L_t$,  parameterized by $\phi^L$, stands for the transformation from a raw data distribution $\mathcal{X}^L$ to a desired distribution $\tilde{\mathcal{X}}^L$ without non-stationary information, and $g^{-1}_{\phi^H}: \bm{x}^H_t \rightarrow \tilde{\bm{x}}^H_t$ parameterized by $\phi^H$ denotes the reverse transformation from $\tilde{\mathcal{X}}^H$ to $\mathcal{X}^H$.
Accordingly, we emit a forecast $\hat{\bm{y}}^H_t$ given an input $\bm{x}^L_t$ as:
\begin{align}
    \hat{\bm{y}}^H_t = g^{-1}_{\phi^H} \left(
        f_\theta \left( g_{\phi^L} \left( \bm{x}^L_t \right)
        \right)
    \right)
    \label{eq:decouple_forecast}
\end{align}

\subsection{Bi-level Optimization for the Decoupled Transformation and Forecasting}
\label{sec:method_optim}

Given the decoupled formulation, we further consider a learning procedure to fulfill the needs of 1) removing distribution shifts that hinder the forecasting performance and 2) performing accurate forecasting based on samples from a transformed stationary distribution.
Motivated by these two-sided objectives, we naturally develop a \textit{bi-level optimization} problem~\cite{colson2007overview}:
\begin{equation}
\begin{aligned}
\phi_* & = \underset{\phi}{\arg \min}\; \mathcal{L} \left( \theta_*(\phi), \phi, \mathcal{T}_{val} \right) \\
\operatorname{s.t.} \quad \theta_*(\phi) &= \underset{\theta}{\arg \min}\; \mathcal{L} \left( \theta, \phi, \mathcal{T}_{train} \right)
\end{aligned}
\label{eq:bilevel_optim}
\end{equation}
where $\phi$ encapsulates $\phi^H$ and $\phi^L$,
$\mathcal{T}_{train}$ and $\mathcal{T}_{val}$ denote the sets of time steps in training and validation data, respectively,
and $\mathcal{L}(\theta, \phi, \mathcal{T}) = \sum_{t \in \mathcal{T}} \ell ( \bm{y}^H_t, g^{-1}_{\phi^H} ( f_\theta ( g_{\phi^L} (\bm{x}^L_t) ) ) )$ is the summation of losses, specified by $\ell (\bm{y}^H_t, \hat{\bm{y}}^H_t)$, over all time steps in $\mathcal{T}$.
The inner loop optimization indicates that given any transformation model parameterized by $\phi$, the solution $\theta_*(\phi)$ minimizes the forecasting losses on the training data, which ensures the effective learning of the forecasting module on the transformed space.
The outer loop optimization aims to identify the transformation solution $\phi_*$ that contributes to the best generalization performance on the validation data, which aligns with our motivation to remove uncertain shifted information hindering forecasting.

Moreover, a standard approach to solve the bi-level optimization problem of Equation (\ref{eq:bilevel_optim}) is to introduce hyper gradients~\cite{chen2022gradient}.
Nevertheless, the calculation of hyper gradients involves the computation of $\frac{\partial \mathcal{L}(\theta_*(\phi), \phi, \mathcal{T}_{val})}{\partial \theta_*(\phi)} \frac{\partial \theta_*(\phi)}{\partial \phi}^T$, which can be prohibitive when performing exact inner optimization.
Therefore, we adopt the first-order approximation, as did in \cite{liu2018darts}, to cut off the gradient dependency introduced by $\frac{\partial \theta_*(\phi)}{\partial \phi}$ and alternate between the following two gradient updates:
\begin{equation}
\begin{aligned}
    \theta_{i+1} &= \theta_i - \alpha \frac{\partial \mathcal{L}(\theta_i, \phi_i, \mathcal{T}_{train})}{\partial \theta_i}, \\
    \phi_{i+1} &= \phi_i - \gamma \frac{\partial \mathcal{L}(\theta_{i+1}, \phi_i, \mathcal{T}_{val})}{\partial \phi_i}.
\end{aligned}
\end{equation}
where $i$ means the $i$-th iteration; $\alpha$ and $\gamma$ are the learning rate for the forecasting module and the transformation module respectively.
In this way, we actually turn the original bi-level optimization into alternate optimization \cite{chen2022gradient}.
While in practice, we find that applying such an approximation to our case brings very fast computation with negligible performance compromise.

\begin{figure}[!t]
\centering
\subfigure[RealNVP]{
\includegraphics[width=0.33\linewidth]{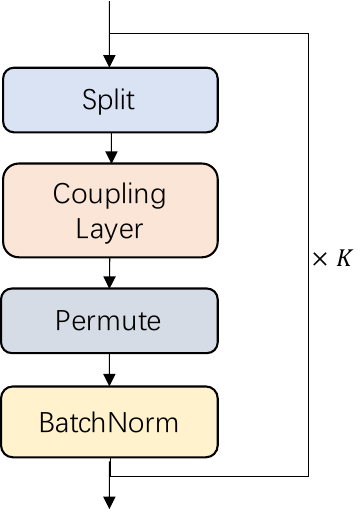}
}
\hspace{-0mm}
\subfigure[IN-Flow]{
\includegraphics[width=0.35\linewidth]{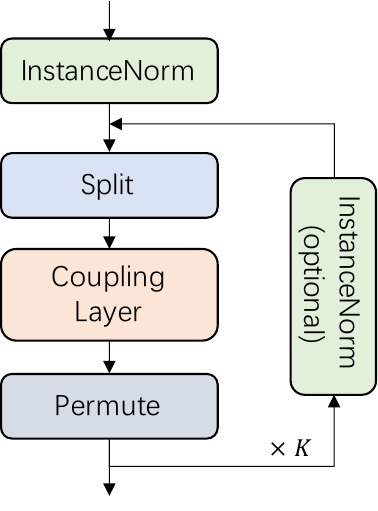}
}
\vspace{-1mm}
\caption{The specific architecture of RealNVP \cite{dinh2016density} and our IN-Flow. We discuss the difference of pre- and post-norm in Section \ref{sec:exp_ablation} and take the pre-norm version as IN-Flow.}
\label{fig:method_inflow}
\vspace{-1mm}
\end{figure}

\begin{figure*}[!h]
\centering
\subfigure[Comparison of forecasting on synthetic data-1]{\includegraphics[width=0.48\linewidth]{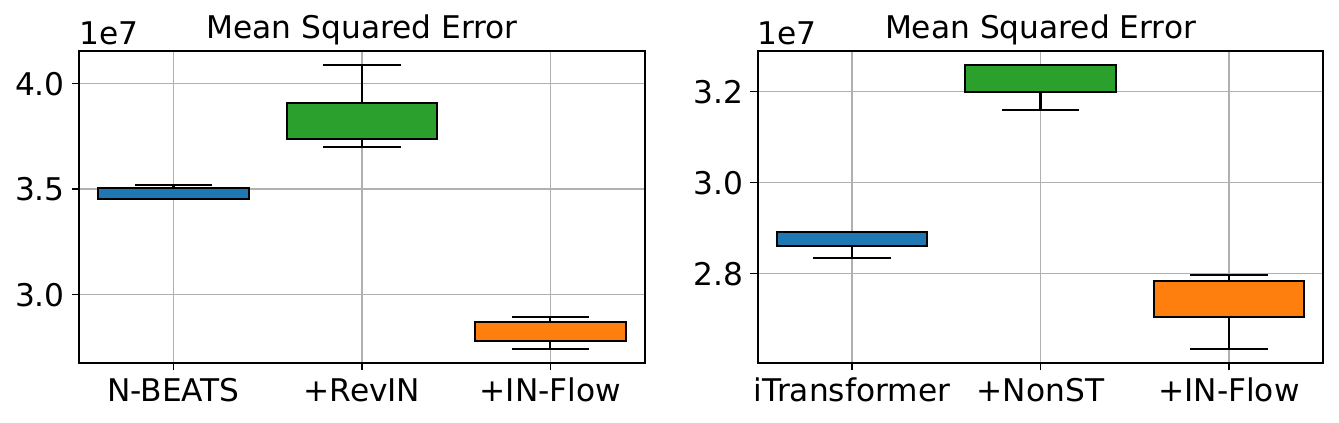}
\label{fig:optimization_case_training_loss}}
\subfigure[Comparison of forecasting on synthetic data-2]{\includegraphics[width=0.48\linewidth]{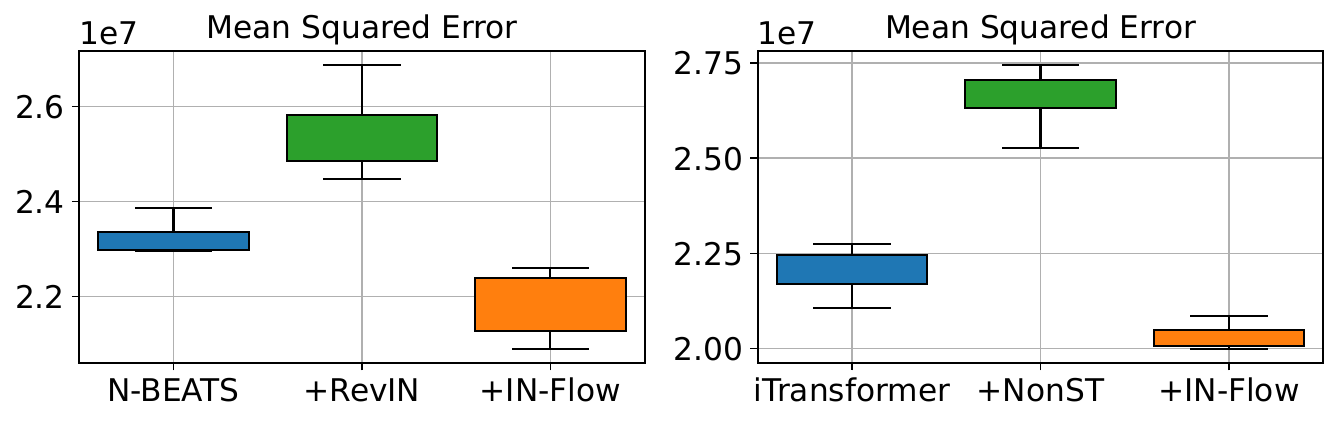}}
\vspace{-3mm}
\caption{Evaluation of forecasting on the synthetic data. NonST means non-stationary transformer.}
\label{fig:main_syn_evaluation}
\vspace{-1mm}
\end{figure*}

\subsection{Instance Normalizaiton Flow} 
\label{sec:method_in_flow}

Though the decoupled formulation with bi-level optimization can accommodate any $f_\theta$, another critical challenge is the design of the transformation module and optimization for $\phi$.
First, the module should have adequate ability to transform data distributions effectively (expressiveness).
Second, the module is responsible for the bi-directional transformation between $\mathcal{X}$ (${\mathcal{X}}^L, {\mathcal{X}}^H$) and $\tilde{\mathcal{X}}$ ($\tilde{\mathcal{X}}^L, \tilde{\mathcal{X}}^H$). Such considerations inspire us to leverage the key idea of \textit{normalizing flows} \cite{kobyzev2020normalizing}, one kind of invertible networks \cite{ardizzone2018analyzing} that transform data distributions.

Existing normalizing flows conduct reversible transformation on one raw data distribution (see Section \ref{sec:bg_flows}). In contrast, we notice our formulation is composed of two transformations, $g_{\phi^L}: \bm{x}_t^L \rightarrow \tilde{\bm{x}}_t^L$ and $g_{\phi^H}^{-1}: \tilde{\bm{x}}_t^H \rightarrow \bm{x}_t^H$, which actually operate in two variable spaces ($g_{\phi^L}$ on $\mathcal{X}^L \in \mathbb{R}^{L \times D}$ and $g_{\phi^H}$ on $\mathcal{X}^H \in \mathbb{R}^{H \times D}$), due to the difference of lookback length ($L$) and horizon length ($H$). 
Nonetheless, these two distributions, $p_{\mathcal{X}^L}(\bm{x}^L)$ and $p_{\mathcal{X}^H}(\bm{x}^H)$, undoubtedly share many common parts in distribution because their samples are generated by segmenting the same time series ($\bm{s}$).
Therefore, to fulfill bi-directional transformations on two variable spaces and capture the common patterns between them, we develop a novel invertible neural architecture, namely \textit{instance normalization flow} (IN-Flow), that instantiates $g_{\phi^L}$ and $g_{\phi^H}$ as a single network and works on variable length of time series, also referred to as $g_{\phi}$ for brevity.

Flow-based models \cite{dinh2014nice,dinh2016density,papamakarios2017masked,kingma2018glow} usually adopt batch normalization in their architectures for a stable training. However, a batch in time series forecasting may consist of samples of various distributions due to distribution shift~\cite{kim2022RevIN}. In such cases, batch normalization could meet with instability in calculating batch statistics and influence the transformation \cite{shen2020powernorm}. Moreover, we notice instance normalization has recently been effective in computer vision \cite{ulyanov2016instance,huang2017arbitrary} and can be an alternative to get over the instability \cite{dumoulin2016learned}. And in the domain of time series, the sequence format can make distribution shift relieved with the application of instance normalization~\cite{kim2022RevIN}.
Inspired by them, our IN-Flow includes parametric instance normalization layer which is written given $d$-th variate by:
\begin{equation}
    {\bm{h}'}^{L,(d)}_{t} = 
    \left( \bm{h}_{t}^{L,(d)}- \mathbf{\mu}_{t}^{(d)} \right)
    {\left( \sigma^{2(d)}_{t} +\epsilon \right)}^{-\frac{1}{2}}
    \operatorname{exp}(\gamma^{(d)}_t)
    + \beta_t^{(d)}
\label{eq:instance_norm_forward}
\end{equation}
where ${\bm{h}'}_t^{L,(d)}$ are the corresponding mappings of $\bm{h}_{t}^{L,(d)} \in \mathbb{R}^{L\times1}$.

For brevity, supposing the current layer is the first layer of $g_\phi$, we have $\bm{h}_{t}^{L,(d)} \triangleq \bm{x}_{t}^{L,(d)}$, 
 we have $\bm{h}_{t}^{L,(d)} \triangleq \bm{x}_{t}^{L,(d)}$,
$\mu_{t}^{(d)} = \frac{1}{L} \sum \bm{x}_{t}^{L,(d)} = \frac{1}{L} \sum_{t-L}^{t-1} s_t^{(d)}$ as the instance mean, and we have $\sigma^{2(d)}_{t} = \frac{1}{L} \sum_{t-L}^{t-1} \left( {s}_{t}^{(d)} - \mu_t^{(d)} \right)^2$ as square of instance standard deviation, where  ${s}_{t}^{(d)}$ as the value of $d$-th variate at timestamp $t$. And $\gamma_t^{(d)}$, $\beta_t^{(d)}$ are affine parameters for transformation in normalization layers.
If it's in the middle layer of IN-Flow, we have $\mu_{t}^{(d)} = \frac{1}{L} \sum \bm{h}_{t}^{L,(d)}$ and $\sigma^{2(d)}_{t} = \frac{1}{L} \sum_{t-L}^{t-1} \left( {h}_{t}^{L,(d)} - \mu_t^{(d)} \right)^2$.
While Equation (\ref{eq:instance_norm_forward}) is the forward transformation of $p(\tilde{\bm{x}}^L_t | \bm{x}^L_t)$, since it is an invertible network, we can also write the inverse transformation of the above operation corresponding for $p(\bm{y}^H_t | \tilde{\bm{y}}^H_t)$ in IN-Flow. Specifically, in the inverse transformation, the instance normalization given the $d$-th variate is by:
\begin{equation}
    {\bm{h}}_{t}^{H,(d)} = \left( {\bm{h}'}^{H,(d)}_{t} - \beta_t^{(d)} \right)
    \operatorname{exp}(-\gamma_t^{(d)})
    {\left( \sigma^{2(d)}_{t} +\epsilon \right)}^{\frac{1}{2}} + \mu_{t}^{(d)}
\end{equation}
where $\bm{h}^{H,(d)}_{t}$ are the corresponding forecasts of $\bm{h}^{L,(d)}_{t}$ in the transformed space; ${\bm{h}}_{t}^{H,(d)}$ are inverse transformed results of current instance normalization layer; other parameters can be  corresponding for those in Equation (\ref{eq:instance_norm_forward}).

Moreover, to make the transformation more expressive, we follow previous models to include 
one common reversible design of flows, \textit{affine coupling layers} \cite{dinh2014nice,dinh2016density} as part of IN-Flow. In implementation, we let the coupling layers work on the feature (variate) dimension by:
\begin{equation}
\left\{\begin{array}{l}
{\bm{h}'}_{t}^{L, (1:d_c)}=\bm{h}_{t}^{L,(1:d_c)}  \\
{\bm{h}'}_{t}^{L,(d_c:D)} = \bm{h}_{t}^{L,(d_c:D)} \odot  s \left( \bm{h}_{t}^{L,(1:d_c)} \right)  + 
  t\left( \bm{h}_{t}^{L,(1:d_c)} \right)
\end{array}\right.
\label{eq:method_coupling_layer}
\end{equation}
where usually we let $d_c = \lceil \frac{D}{2} \rceil$ stand for the split position; $\bm{h}_{t}^{L,(1:d_c)} \in \mathbb{R}^{L \times d_c}$ denotes data composed of first $d_c$ variates of $\bm{h}_{t}^{L}$ and $\bm{h}_{t}^{L,(d_c:D)} \in \mathbb{R}^{L \times (D-d_c)}$ denotes the left $D-d_c$ variates; $s$ and $t$ are scale and translation functions mapping from $\mathbb{R}^{d_c}$ to $\mathbb{R}^{D-d_c}$; ${\bm{h}'}_{t}^{L}$ is the processed output. Accordingly, the inverse transformation of the above coupling layers for $p(\bm{y}^H_t | \tilde{\bm{y}}^H_t)$ in IN-Flow can be written by:
\begin{equation}
\left\{\begin{array}{l}
{\bm{h}}_{t}^{H, (1:d_c)}={\bm{h}'}_{t}^{H,(1:d_c)}  \\
{\bm{h}}_{t}^{H,(d_c:D)} = \left( {\bm{h}'}_{t}^{H,(d_c:D)}  -
  t\left( {\bm{h}'}_{t}^{H,(1:d_c)} \right) \right)
/ s \left( {\bm{h}'}_{t}^{H,(1:d_c)} \right)
\end{array}\right.
\label{eq:appendix_coupling_layer_backward}
\end{equation}
where $ {\bm{h}'}_{t}^{H}$ are the forecasting results of ${\bm{h}'}_{t}^{L}$ in the hidden space; $ {\bm{h}}_{t}^{H}$ is the processed results of $ {\bm{h}'}_{t}^{H}$ by the affine coupling layers.
Especially, if the coupling layers are the final layers of IN-Flow, we can have $\hat{\bm{y}}_{t}^{H}={\bm{h}}_{t}^{H}$.
In stacking IN-Flow layers, we find that the pre-norm (IN-Flow) can achieve better performance than post-norm (IN-Flow-T in Section \ref{sec:exp_ablation}). Thus we take the pre-norm variant as the final design of IN-Flow.

Overall, the main architecture of IN-Flow is by stacking \textit{instance normalization} and \textit{coupling layers} iteratively, similar to other classic normalizing flows such as RealNVP \cite{dinh2016density}, as shown in Figure \ref{fig:method_inflow}.
We also adopt \textit{Split} and \textit{Permute} operations \cite{dinh2014nice,dinh2016density,kingma2018glow}
to assist for stacking layers. 
Note that both Equation (\ref{eq:instance_norm_forward}) and Equation (\ref{eq:method_coupling_layer}) belong to the transformation of $p(\tilde{\bm{x}}^L_t | \bm{x}^L_t)$. We can also accordingly write the \textit{inverse} transformation of above layers for $p(\bm{y}^H_t | \tilde{\bm{y}}^H_t)$ in IN-Flow. 

%% file: 5_experiments.tex

\section{Experiments}
\label{sec:experiment}
Our empirical studies in this section aim to answer the following questions: 
1) With some solutions existing, why should we further study distribution shifts and non-stationary time series forecasting?  
2) How much benefit can our framework gain for time series forecasting compared with existing state-of-the-art models in real-world settings? 
3) Does each part of {IN-Flow} designs (including optimization) count for good forecasting performance?

\noindent \textbf{Baselines.} We mainly consider several kinds of state-of-the-art methods for non-stationary time series forecasting. For the model-agnostic normalization technique, we adopt RevIN \cite{kim2022RevIN}, Dish-TS~\cite{fan2023dish}, and SAN~\cite{liu2023adaptive}. For the backbones, we adopt the recently well-performed forecasting methods, i.e., iTransformer~\cite{itransformer_2024}, PatchTST~\cite{patchtst_2023}, Autoformer~\cite{wu2021autoformer}, and N-BEATS~\cite{Oreshkin2020N-BEATS}. 
For the model-specific mechanisms, we adopt Non-tationary Transformer~\cite{liu2022NonstatTranformers} and we take vanilla Transformer \cite{vaswani2017attention}, Informer \cite{zhou2021informer} and Autoformer \cite{wu2021autoformer} as the backbones
More baseline details are in Appendix \ref{sec:appendix_baseline}.


\noindent \textbf{Evaluation Details.}
We evaluate time series forecasting performance on the classic mean squared error (MSE) and mean absolute error (MAE). To directly reflect distribution shifts in time series, all the evaluations are conducted on original data without data scaling or normalization following settings of previous works~\cite{fan2023dish}.
The reported metrics on real-world data are scaled for readability. More evaluation details are included in Appendix \ref{sec:appendix_more_real_evaluation_details}.

\subsection{Evaluation on Synthetic Data} \label{sec:exp_synthetic}
To intuitively illustrate the importance of modeling distribution shifts in forecasting and the superiority of IN-Flow, we generate synthetic time series data with distributions shifted. We first group each neighbored $\tau$ points as a segment and let every segment follows one distribution. Supposing to synthesize one series $\{s_1, \cdots, s_T\}$ of length $T$, this series can be regarded to have $U = \lceil T/\tau \rceil$ segments, where the $u$-th segment follows its distribution $\mathcal{P}_u$. 
We can thus control the degree of time series distribution shift through adjusting $\tau$ and $\mathcal{P}_u$. Under such settings, both $p(\bm{x}_t^{L})$ and $p( \bm{y}^{H}_t|\bm{x}_t^{L})$ keep shifted over time if lookback length $L>\tau$ and lookahead length $H>\tau$. Then for each $\mathcal{P}_u$, we simply make a regular cosine wave signal to model simple periodic patterns in time series. Specifically, we let $s_t =  A_u \operatorname{cos}(2\pi\frac{1}{T_u} t+B_u) + C_u$ if time index $t$ belongs to $u$-th segment, where $A_u, T_u, B_u, C_u$ indicate amplitude, period, phase and level parameters for the $u$-th segment. 

\input{tables/main_exp.tex}

Following the above rules, we make three synthetic datasets, called Synthetic-1/2 by varying different $\tau$ and $\mathcal{P}_u$. 
Due to space limit, we include more details about synthetic datasets in Appendix \ref{sec:appendix_more_syn}. 
Figure \ref{fig:main_syn_evaluation} shows two different comparisons of different backbones on Synthetic-1 and Synthetic-2 datasets. We notice that in our simulation settings, both two state-of-the-art techniques RevIN and non-stationary Transformers would fail significantly, causing a more than 10\% performance decrease on iTransformer and N-BEATS respectively. These results demonstrate their limitations in overcoming distribution shifts in forecasting when meeting some severely shifted time series, while IN-Flow can still improve the forecasting ability of backbones by a large margin which shows our superiority in non-stationary forecasting.

\vspace{-1mm}
\subsection{Evaluation on Real-world Data} \label{sec:exp_real_world}
Apart from simulation experiments, we further perform experiments on real-world datasets. We adopt six existing shifted datasets following previous forecasting literature: Electricity transformer temperature datasets 
include time series of oil temperature and power load collected from electricity transformers in China. \textbf{\textit{ETTm2}} dataset is recorded at a fifteen-minute frequency ranging from July 2016 to July 2018.
In contrast, \textbf{\textit{ETTh1}} dataset dataset is recorded every hour.
\textbf{\textit{Electricity}} dataset contains the hourly electricity
consumption of 321 customers from 2012 to 2014.
\textbf{\textit{Weather}} dataset 
contains meteorological measurements of 21 weather indicators, which are collected 10-minutely in 2020.
\textbf{\textit{CAISO}} dataset 
includes hourly electricity load series in different zones of California ranging from 2017 to 2020. \textbf{\textit{NordPool}} dataset 
includes hourly energy production volume series of several European countries in 2020. 

For a stable evaluation, we perform z-score normalization for each backbone model and inverse the forecasting results to report metrics.
We rerun all the models under four different seeds to report the average MSE/MAE on the testset of each dataset. 
For data split, we follow \cite{zhou2021informer} and split data into train/validation/test set by the ratio 6:2:2 towards \textit{ETTh1} dataset \textit{ETTm2} dataset. We also adopt the ratio 6:2:2 to split the data for \textit{CAISO} dataset and \textit{NordPool} dataset. For \textit{Weather} dataset and Electricity dataset, we follow \cite{wu2021autoformer} to split data by the ratio of 7:1:2 for the train/validation/test set. 
To cover short/long-term forecasting, we set the length of the lookback and horizon window from 48 to 336. We train the forecasting models using L2 loss and Adam \cite{kingma2014adam} on a single NVIDIA A100 40GB GPU. 

\subsubsection{Main Results}

Table \ref{table:main_exp} demonstrates the overall performance comparison of the basic models and their IN-Flow-equipped versions on four representative time series forecasting models. 
Based on the results, we can observe that IN-Flow consistently improves backbone models under different settings across the datasets. The average improvements of IN-Flow on the backbone models come to 28.3\% on PatchTST, 21.3\% on iTransformer, 28.5 on Autoformer, and 12.1\% on N-BEATS across different datasets, which can show the effectiveness of our methods. Moreover, in certain situations, the improvements are even higher.
Notably, with the lookback/horizon length as 168 on the Weather dataset, IN-Flow improves iTransformer by 32.5\% ($0.710 \to 0.479$) and improves PatchTST by 33.2\% ($0.882 \to 0.589$) on MSE.
Another case on CAISO dataset is that the MSE of Autoformer with IN-Flow is reduced from 1.789 to 0.823, which achieves a very large (54.0\%) improvement. The overall superior performance shows the great potential of IN-Flow in enhancing deep forecasting models on prediction.


\vspace{-1.4mm}

\subsubsection{Comparison with Other Non-stationary Forecasting Techniques}

In this section, we further compare our performance with more recent normalization techniques, Dish-TS \cite{fan2023dish} and SAN~\cite{liu2023adaptive} that handle distribution shifts in time series forecasting. Table \ref{table:results_with_norm} has shown the performance comparison in time series forecasting taking the PatchTST~\cite{patchtst_2023} as the backbone. From the results, we can observe that the existing  Dish-TS can only improve the backbone in some shifted datasets. In some situations, Dish-TS might lead to worse performances, while SAN can perform better than it. Nevertheless, our IN-Flow can usually achieve the best  performance. A potential explanation is that IN-Flow transforms data with multiple coupling and normalization layers with much expressiveness.

Besides, we also include the performance of IN-Flow in Table \ref{table:main_nonstation} compared with the SOTA architecture-specific method, Non-stationary Transformers \cite{liu2022NonstatTranformers}, in which we consider three transformer-based models as backbones to compose Transformer, Informer, and Autoformer. We can easily observe that IN-Flow achieves a significant improvement over Non-stationary Transformers. 
The superiority signifies that even though IN-Flow is a model-agnostic method, it can have great adaptation for transformer-based models and even beat architecture-specific methods with tailor-designed mechanisms de-stationary attention in \cite{liu2022NonstatTranformers}, which reveals the great potential of its compatibility to deep forecasting models.

\input{tables/main_non_station_dish}

\input{tables/main_nonstaion.tex}

\subsection{Additional Experiments}


\subsubsection{Ablation Studies} \label{sec:exp_ablation}

In order to verify the effectiveness of our designs, we conduct adequate ablation studies by adjusting components of IN-Flow and considering some other transformation methods. Specifically, we consider four different variants for ablation tests:
\textbf{RealNVP} \cite{dinh2016density} adopts the coupling layers \cite{dinh2014nice} and the batch normalization \cite{ioffe2015batch}. In implementation, we mainly follow the reproduced implementation\footnote{\url{https://github.com/kamenbliznashki/normalizing_flows}} and abandon some complicated modules such as multi-scale architectures \cite{dinh2016density,kingma2018glow}. This method is shown in Figure \ref{fig:appendix_subfig_realnvp} of Appendix.
\textbf{RealNVP-c} \cite{dinh2016density} is a variant of RealNVP which only adopts the coupling layers. We use this method as comparison to verify the effectiveness of batch normalization \cite{ioffe2015batch} for transformation in forecasting. The method is shown in Figure \ref{fig:appendix_subfig_realnvp-c} in Appendix.
\textbf{IN-Flow-J} is a variant of IN-Flow (as shown in Figure \ref{fig:appendix_subfig_Inflow} in Appendix), which jointly optimizes the transformation module and the forecasting module on the training data without 
    bi-level optimization in the formulation. This method is to verify the effectiveness of bi-level optimization in IN-Flow.
\textbf{IN-Flow-T} (post-norm) is another variant of IN-Flow, which changes the relative position of coupling layers and instance normalization and puts the instance normalization to the {t}ail of flows, as shown in Figure \ref{fig:appendix_subfig_Inflow-T} in Appendix. This method is to verify the architecture design of IN-Flow.
    
We conduct extensive ablation studies considering above four kinds of variants and taking N-BEATS and Autoformer as the backbone models respectively.
We include the experimental ablation results on N-BEATS in Table \ref{table:appendix_ablation_nbeats}. 
Based on the results, we can notice that RealNVP (with batch normalization) can always perform the worst in the comparison. This signifies the batch normalization is not proper for the transformation of time series. However, with pure coupling layers (RealNVP-c), it can perform well in some cases.
Both of these two observations motivate us to remove batch normalization for transformation and further propose the instance normalization  instead. 
We can easily observe that, IN-Flow and its variants can always beat RealNVP series methods, which demonstrates the superiority of the instance normalization. 
Moreover, we notice that the joint optimization can sometime work well, especially in CAISO dataset. However, the overall performance of bi-level optimization becomes more stable and competitive. Thus, we finally adopt the bi-level optimization in our proposed methods.

\input{tables/main_ablation_cr}

\begin{figure}[!t]
\centering
\subfigure[Training Loss]{
\includegraphics[width=0.49\linewidth]{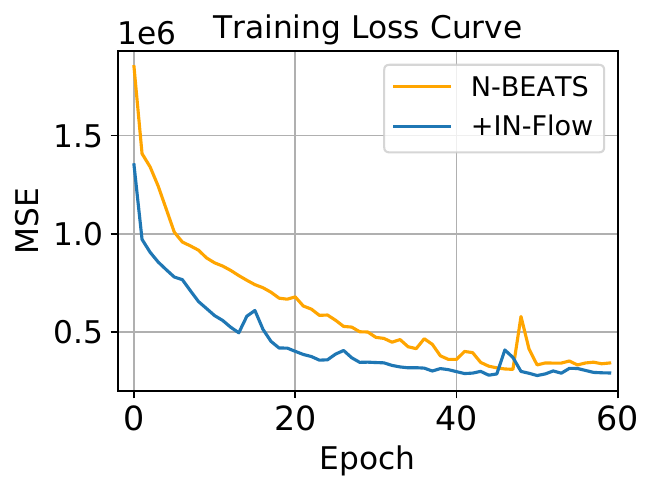}
\label{fig:optimization_case_training_loss}
}
\hspace{-4mm}
\subfigure[Validation Loss]{
\includegraphics[width=0.468\linewidth]{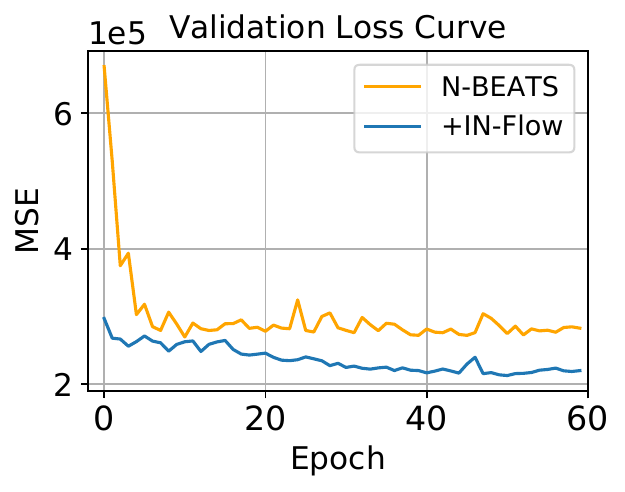}
\label{fig:optimization_case_validation_loss}
}
\vspace{-4mm}
\caption{Learning loss curves with and without IN-Flow.}
\vspace{-2mm}
\end{figure}

\subsubsection{Training Analysis.} To further study the learning process of IN-Flow, we visualize the learning loss curves of N-BEATS and IN-Flow. Figure \ref{fig:optimization_case_training_loss} and Figure \ref{fig:optimization_case_validation_loss} demonstrate the training loss and validation loss curve on CAISO dataset respectively. 
We notice that with IN-Flow the training loss of N-BEATS has faster convergence 
which signifies the transformation module improves the learning of the forecasting module. After about 50 epochs, though the two training loss curves come close, the validation loss of IN-Flow is far lower than pure N-BEATS, which is attributed to the transformation module that relieves distribution shifts for fewer validation errors and better generalization in forecasting.

\begin{figure*}[!t]
\centering
\hspace{+1mm}
\subfigure[RevIN]{\includegraphics[width=0.23\linewidth]{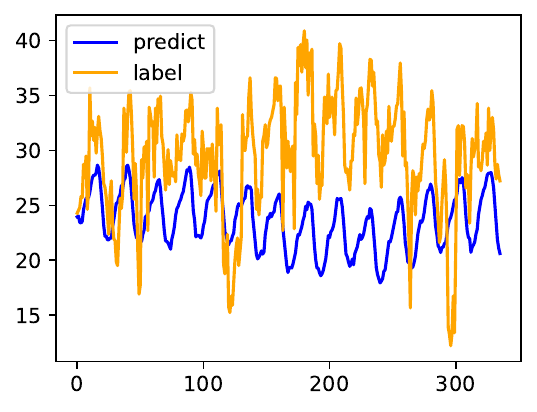}
\label{fig:optimization_case_training_loss}}
\hspace{+1mm}
\subfigure[Dish-TS]{\includegraphics[width=0.23\linewidth]{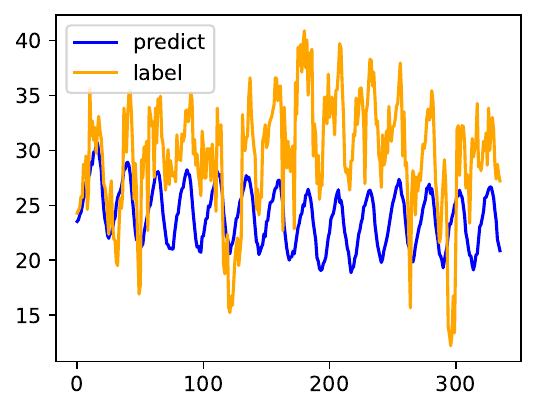}
\label{fig:optimization_case_validation_loss}}
\hspace{+1mm}
\subfigure[SAN]{\includegraphics[width=0.23\linewidth]{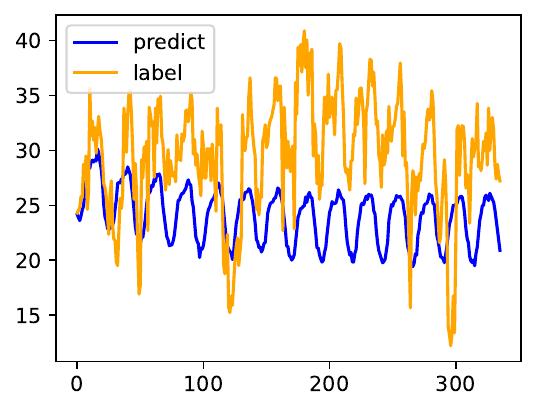}
\label{fig:optimization_case_validation_loss}}
\hspace{+1mm}
\subfigure[IN-Flow]{\includegraphics[width=0.23\linewidth]{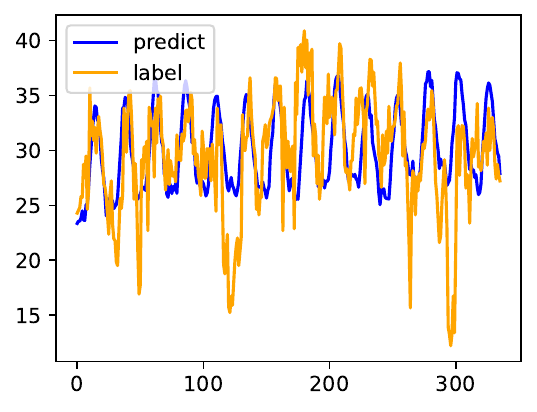}
\label{fig:optimization_case_validation_loss}}
\vspace{-2mm}
\caption{Visualization of long-term 336 steps forecasting results of a test sample in ETTm2 dataset.} \label{fig:case_study}
\vspace{-1mm}
\end{figure*}

\subsubsection{Case Visualization}
We also include visualization case study towards forecasting. Figure \ref{fig:case_study} has shown a test sample predicted by RevIN, Dish-TS, SAN, and IN-Flow in the ETTm2 dataset when taking PatchTST as the backbone. Based on the results, we can find that all four methods can capture certain seasonal patterns. However, the amplitudes of the seasonal patterns are usually correctly learned. More importantly, we notice that in both RevIN, Dish-TS and SAN, the trend patterns are not correctly modeled, leading to a larger deviation towards the ground truth. This signifies that existing normalization techniques might abandon some useful information and thus have some worse cases. In contrast, our IN-Flow has adequate transformation ability for finishing forecasting by learning both trend and seasonality accurately.

\subsubsection{Distribution Visualization} \label{sec:distribution}
\begin{figure}[!t]
\centering
\subfigure[Raw distribution]{\includegraphics[width=0.48\linewidth]{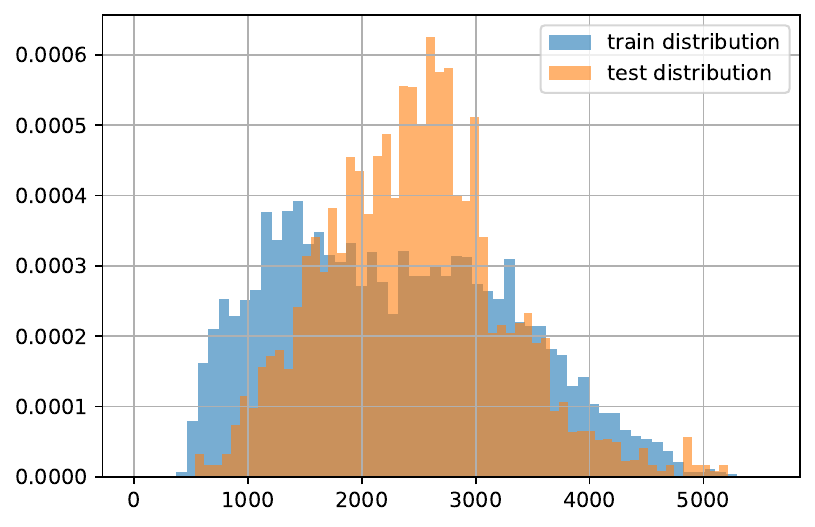}
}
\subfigure[IN-Flow transformed distribution]{\includegraphics[width=0.465\linewidth]{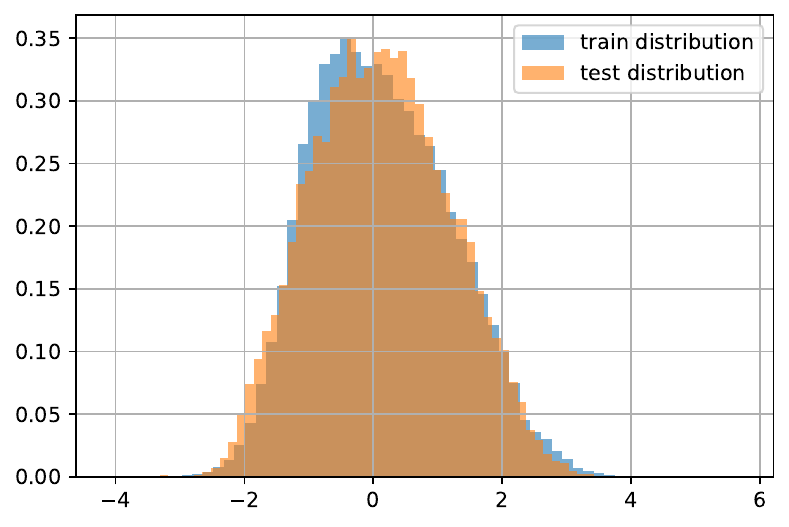}
}
\vspace{-2mm}
\caption{The distribution visualization on NordPool dataset.} \label{fig:dist}
\vspace{-1mm}
\end{figure}

In order to intuitively show the effectiveness of our IN-Flow, we have visualized the distribution of train and test sets before and after the transformation. Since the original time series is not i.i.d. and the window splitting decides the distribution, we visualize the distribution of the mean value of windows. Figure \ref{fig:dist} shows the comparison of the distribution of train and test sets on the NordPool dataset. We can easily observe that in the original data, the train and test sets have obvious differences. After being transformed by IN-Flow, the distribution is much less distinguishable, showing the shift alleviation function of IN-Flow.

\subsubsection{Statistic Measurements} \label{sec:distribution_stats}
We also include the statistical measurements of the distribution difference of train and test distribution before or after the transformation of IN-Flow. We select 
Wasserstein distance as our measurements. In order to intuitively show the effectiveness of our IN-Flow, we have visualized the distribution of train and test sets before and after the transformation. Since the original time series is not i.i.d. and the window splitting decides the distribution, we visualize the distribution of the mean value of windows of the data.
The distribution difference between the train and test sets is shown in Table \ref{tab:wd}. 
From the results, we can observe that IN-Flow can largely reduce the distribution difference between train and test distribution. This provides large potential for the non-stationary forecasting.

\begin{table}[!h]
\centering
\small
\caption{Wassertein distance of train and test set.} \label{tab:wd}
\vspace{-2mm}
\resizebox{0.86\linewidth}{!}{
\begin{tabular}{llllllll}
\toprule
Dataset                     & {\scriptsize Synthetic-2}    & {\scriptsize ETTh1}  & {\scriptsize Electricity}   \\
\midrule
Train/Test Raw WD           & 0.2213  & 0.3270 & 0.3337    \\
WD (with IN-Flow) & 0.0783  & 0.0702 & 0.0547    \\
\midrule
Dataset  & {\scriptsize Weather} & {\scriptsize CAISO}  & {\scriptsize NordPool} \\
\midrule
Train/Test Raw WD & 0.1751  & 0.0944 & 0.1170\\
WD (with IN-Flow) & 0.0775  & 0.0857 & 0.0471\\
\bottomrule
\end{tabular}
}
\vspace{-1mm}
\end{table}

%% file: tables/main_exp.tex
\begin{table*}[ht]
\centering
\resizebox{\linewidth}{!}{
\begin{threeparttable}
\small
\caption{Performance comparison of time series forecasting on backbone models and IN-Flow. The length of lookback/horizon windows is prolonged from 48 to 336 to cover short and long forecasting settings.
} \label{table:main_exp}
\vspace{-4mm}
\begin{tabular}{c|c|cccc|cccc|cccc|cccc}
\toprule
\multicolumn{2}{c}{Method}& \multicolumn{2}{c}{PatchTST}&\multicolumn{2}{c}{+{IN-Flow}}& \multicolumn{2}{c}{iTransformer}&\multicolumn{2}{c}{{+IN-Flow{}}}& 
\multicolumn{2}{c}{{Autoformer}}&   \multicolumn{2}{c}{+{IN-Flow}} & \multicolumn{2}{c}{N-BEATS}&\multicolumn{2}{c}{+{IN-Flow}} \\ 
\cmidrule(r){1-2} \cmidrule(r){3-4} \cmidrule(r){5-6} \cmidrule(r){7-8} \cmidrule(r){9-10} \cmidrule(r){11-12} \cmidrule(r){13-14} \cmidrule(r){15-16} \cmidrule(r){17-18} 
\multicolumn{2}{c}{Metric}&
MSE& MAE&  MSE& MAE&
MSE& MAE&  MSE& MAE&
MSE& MAE&  MSE& MAE&
MSE& MAE&  MSE& MAE\\
\midrule \midrule
\multirow{4}{*}{ \rotatebox{90}{ETTh1} }
&48  & 1.128 & 1.865 & \textbf{0.936} & \textbf{1.635} & 1.051 & 1.717 & \textbf{0.811} & \textbf{1.244} & 1.237 & 2.010  & \textbf{1.215} & \textbf{1.937} & 0.959 & 1.637 & \textbf{0.951} & \textbf{1.624} \\
&96  & 1.320  & 2.109 & \textbf{1.025} & \textbf{1.782} & 1.188 & 1.933 & \textbf{0.896} & \textbf{1.542} & 1.710  & 2.347 & \textbf{1.428} & \textbf{2.141} & 1.072 & 1.789 & \textbf{1.027} & \textbf{1.768} \\
&168 & 2.001 & 2.679 & \textbf{1.077} & \textbf{1.888} & 1.917 & 2.624 & \textbf{0.901} & \textbf{1.925} & 2.139 & 2.739 & \textbf{1.303} & 
\textbf{2.095} & 1.223 & 1.989 & \textbf{1.139} & \textbf{1.914} \\
&336 & 1.784 & 2.613 & \textbf{1.121} & \textbf{2.021} & 1.734 & 2.568 & \textbf{1.107} & \textbf{1.189} & 2.444 & 2.965 & \textbf{1.502} & \textbf{2.305} & 1.222 & 2.083 & \textbf{1.146} & \textbf{2.044} \\
\midrule
\multirow{4}{*}{ \rotatebox{90}{ETTm2} }
&48  & 1.116 & 2.204 & \textbf{0.925} & \textbf{1.908} & 1.094 & 2.168 & \textbf{0.958} & \textbf{1.972} & 1.477 & 2.469 & \textbf{1.459} & \textbf{2.432} & 1.329 & 2.308 & \textbf{1.229} & \textbf{2.182} \\
&96  & 1.503 & 2.699 & \textbf{0.992} & \textbf{1.965} & 1.470 & 2.178 & \textbf{0.996} & \textbf{1.973} & 1.962 & 2.901 & \textbf{1.869} & \textbf{2.748} & 1.628 & 2.604 & \textbf{1.570} & \textbf{2.496} \\
&168 & 1.708 & 2.783 & \textbf{1.213} & \textbf{2.202} & 1.352 & 2.472 & \textbf{1.248} & \textbf{2.275} & 1.898 & 2.919 & \textbf{1.834} & \textbf{2.831} & 2.148 & 3.021 & \textbf{1.752} & \textbf{2.681} \\
&336 & 2.664 & 3.465 & \textbf{1.564} & \textbf{2.482} & 2.124 & 2.713 & \textbf{1.678} & \textbf{2.532} & 2.271 & 3.235 & \textbf{2.113} & \textbf{3.003} & 2.376 & 3.339 & \textbf{1.928} & \textbf{2.847} \\
\midrule
\multirow{4}{*}{ \rotatebox{90}{Electricity}}
&48  & 0.632 & 0.366 & \textbf{0.554} & \textbf{0.342} & 0.513 & 0.319 & \textbf{0.421} & \textbf{0.312} & 0.686 & 0.398 & \textbf{0.592} & \textbf{0.381} & 0.698 & 0.431 & \textbf{0.594} & \textbf{0.422} \\
&96  & 0.882 & 0.698 & \textbf{0.589} & \textbf{0.341} & 0.710 & 0.558 & \textbf{0.479} & \textbf{0.313} & 0.936 & 0.763 & \textbf{0.669} & \textbf{0.397} & 0.912 & 0.751 & \textbf{0.872} & \textbf{0.723} \\
&168 & 1.678 & 0.541 & \textbf{0.882} & \textbf{0.389} & 1.341 & 0.413 & \textbf{0.790} & \textbf{0.329} & 1.989 & 0.576 & \textbf{0.943} & \textbf{0.437} & 2.031 & 0.582 & \textbf{1.521} & \textbf{0.503} \\
&336 & 2.482 & 0.814 & \textbf{2.104} & \textbf{0.768} & 2.012 & 0.654 & \textbf{1.678} & \textbf{0.594} & 2.934 & 0.898 & \textbf{2.521} & \textbf{0.813} & 2.835 & 0.872 & \textbf{2.671} & \textbf{0.852} \\
\midrule
\multirow{4}{*}{ \rotatebox{90}{Weather}}
&48  & 1.003 & 3.291 & \textbf{0.413} & \textbf{1.710} & 1.001 & 3.288 & \textbf{0.447} & \textbf{1.897} & 1.297 & 3.217 & \textbf{0.402} & \textbf{1.844} & 0.981 & 3.226 & \textbf{0.432} & \textbf{1.698} \\
&96  & 0.335 & 1.886 & \textbf{0.279} & \textbf{1.541} & 0.334 & 1.886 & \textbf{0.301} & \textbf{1.629} & 0.658 & 3.327 & \textbf{0.285} & \textbf{1.663} & 0.246 & 1.642 & \textbf{0.223} & \textbf{1.328} \\
&168 & 0.338 & 2.138 & \textbf{0.207} & \textbf{1.421} & 0.337 & 2.138 & \textbf{0.263} & \textbf{1.581} & 0.398 & 2.260 & \textbf{0.253} & \textbf{1.708} & 0.192 & 1.392 & \textbf{0.182} & \textbf{1.217} \\
&336 & 0.385 & 2.375 & \textbf{0.192} & \textbf{1.475} & 0.384 & 2.373 & \textbf{0.228} & \textbf{1.535} & 0.570 & 2.996 & \textbf{0.255} & \textbf{1.712} & 0.194 & 1.520 & \textbf{0.179} & \textbf{1.311} \\
\midrule
\multirow{4}{*}{ \rotatebox{90}{CAISO}}
&48  & 0.986 & 0.499 & \textbf{0.641} & \textbf{0.383} & 0.906 & 0.425 & \textbf{0.627} & \textbf{0.377} & 1.789 & 0.740 & \textbf{0.823} & \textbf{0.467} & 1.099 & 0.684 & \textbf{0.514} & \textbf{0.341} \\
&96  & 0.898 & 0.451 & \textbf{0.828} & \textbf{0.422} & 0.833 & 0.424 & \textbf{0.799} & \textbf{0.393} & 2.428 & 0.826 & \textbf{1.289} & \textbf{0.576} & 0.886 & 0.440 & \textbf{0.878} & \textbf{0.428} \\
&168 & 1.369 & 0.577 & \textbf{1.161} & \textbf{0.500} & 1.297 & 0.503 & \textbf{1.071} & \textbf{0.411} & 2.832 & 0.907 & \textbf{1.344} & \textbf{0.575} & 1.189 & 0.504 & \textbf{1.150} & \textbf{0.498} \\
&336 & 2.541 & 0.852 & \textbf{1.636} & \textbf{0.604} & 2.124 & 0.712 & \textbf{1.524} & \textbf{0.456} & 3.148 & 0.960 & \textbf{1.809} & \textbf{0.679} & 1.612 & 0.607 & \textbf{1.518} & \textbf{0.602} \\
\midrule
\multirow{4}{*}{ \rotatebox{90}{NordPool}}
&48  & 1.304 & 0.664 & \textbf{1.156} & \textbf{0.636} & 1.331 & 0.800 & \textbf{1.233} & \textbf{0.769} & 1.632 & 0.765 & \textbf{1.478} & \textbf{0.732} & 1.192 & 0.639 & \textbf{1.150} & \textbf{0.632} \\
&96  & 1.771 & 0.799 & \textbf{1.687} & \textbf{0.777} & 1.851 & 0.936 & \textbf{1.796} & \textbf{0.823} & 2.187 & 0.889 & \textbf{2.071} & \textbf{0.852} & 1.644 & 0.764 & \textbf{1.596} & \textbf{0.756} \\
&168 & 2.513 & 0.986 & \textbf{2.009} & \textbf{0.865} & 2.597 & 1.035 & \textbf{2.584} & \textbf{0.908} & 2.596 & 0.992 & \textbf{2.430} & \textbf{0.957} & 2.044 & 0.867 & \textbf{1.918} & \textbf{0.836} \\
&336 & 2.946 & 1.040 & \textbf{2.010} & \textbf{0.887} & 3.022 & 0.976 & \textbf{2.138} & \textbf{0.941} & 2.993 & 1.236 & \textbf{2.079} & \textbf{0.887} & 2.557 & 0.953 & \textbf{2.098} & \textbf{0.884} \\
\midrule
\bottomrule
\end{tabular}
\vspace{-0.5mm}
\end{threeparttable}
}
\end{table*}

%% file: tables/main_non_station_dish.tex
\begin{table}[!t]
\centering
\caption{Performance comparison of Dish-TS, SAN, and IN-Flow on six datasets under three seeds when taking the PatchTST as the backbone model.} \label{table:results_with_norm}
\vspace{-2mm}
\resizebox{1.0\linewidth}{!}{
\begin{threeparttable}
\small
\begin{tabular}{c|c|cccccccc}
\toprule
\multicolumn{2}{c}{Method}& 
\multicolumn{2}{c}{ PatchTST }&\multicolumn{2}{c}{+Dish-TS}& 
\multicolumn{2}{c}{ +SAN}&\multicolumn{2}{c}{ +IN-Flow}\\
\cmidrule(r){1-2} \cmidrule(r){3-4} \cmidrule(r){5-6} \cmidrule(r){7-8} \cmidrule(r){9-10} 
\multicolumn{2}{c}{Metric}&
MSE& MAE&  MSE& MAE&
MSE& MAE&  MSE& MAE\\
\midrule \midrule
\multirow{4}{*}{ \rotatebox{90}{ETTh1} }
& 48     & 1.128    & 1.865 & 0.975 & 1.661 & 0.971   & 1.668 & \textbf{0.936} & \textbf{1.635} \\
         & 96     & 1.320    & 2.109 & 1.100 & 1.933 & 1.093   & 1.839 & \textbf{1.025} & \textbf{1.782} \\
         & 168    & 2.001    & 2.679 & 1.229 & 2.024 & 1.110   & 1.950 & \textbf{1.077} & \textbf{1.888} \\
         & 336    & 1.784    & 2.613 & 1.352 & 2.172 & 1.251   & 2.137 & \textbf{1.121} & \textbf{2.021} \\

\midrule
\multirow{4}{*}{ \rotatebox{90}{Weather}}
& 48     & 1.003    & 3.291 & 0.852 & 2.288 & 0.466   & 1.753 & \textbf{0.413} & \textbf{1.710} \\
         & 96     & 0.335    & 1.886 & 0.314 & 1.856 & 0.309   & 1.777 & \textbf{0.279} & \textbf{1.541} \\
         & 168    & 0.338    & 2.138 & 0.219 & 1.975 & 0.217   & 1.622 & \textbf{0.207} & \textbf{1.421} \\
         & 336    & 0.385    & 2.375 & 0.256 & 2.612 & 0.239   & 1.907 & \textbf{0.192} & \textbf{1.475} \\

\midrule
\multirow{4}{*}{ \rotatebox{90}{CAISO}}
& 48     & 0.986    & 0.499 & 0.838 & 0.456 & 0.748   & 0.425 & \textbf{0.641} & \textbf{0.383} \\
         & 96     & 1.098    & 0.451 & 1.216 & 0.541 & 0.901   & 0.446 & \textbf{0.828} & \textbf{0.422} \\
         & 168    & 1.369    & 0.577 & 1.396 & 0.558 & 1.216   & 0.527 & \textbf{1.161} & \textbf{0.500} \\
         & 336    & 2.541    & 0.852 & 1.925 & 0.708 & 1.831   & 0.683 & \textbf{1.636} & \textbf{0.604} \\

\midrule
\multirow{4}{*}{ \rotatebox{90}{NordPool}}
& 48     & 1.304    & 0.664 & 1.293 & 0.654 & 1.200   & 0.646 & \textbf{1.156} & \textbf{0.636} \\
         & 96     & 1.971    & 0.892 & 1.895 & 0.924 & 1.812   & 0.800 & \textbf{1.687} & \textbf{0.777} \\
         & 168    & 2.513    & 0.986 & 2.321 & 0.928 & 2.047   & 0.885 & \textbf{2.009} & \textbf{0.865} \\
         & 336    & 2.946    & 1.040 & 2.707 & 1.010 & 2.014   & 0.889 & \textbf{2.010} & \textbf{0.887} \\
\midrule
\bottomrule
\end{tabular}
\end{threeparttable}
}
\vspace{-1mm}
\end{table}

%% file: tables/main_nonstaion.tex
\begin{table}[!h]
\small
\centering
\caption{Performance comparison with non-stationary transformers, where Transformer, Informer, and Autoformer are coupled into non-stationary versions to compare with
IN-Flow-coupled versions. 
} \label{table:main_nonstation}
\vspace{-2mm}
\resizebox{\linewidth}{!}{
\begin{threeparttable}
\begin{tabular}{l|cc|cc|cc}
\toprule
Datasets & \multicolumn{2}{c}{ETTm2} & \multicolumn{2}{c}{Weather} & \multicolumn{2}{c}{CAISO}  \\
Length & 48  & 336 & 48  & 336 & 48 & 336  \\
\midrule
NS-Transformer & 1.677  & 2.983 & 0.551  & 0.323 &
0.850  & 2.282\\
INF-Transformer& \textbf{1.352} & \textbf{2.053} &  \textbf{0.360}   & \textbf{0.210} & \textbf{0.757}  & \textbf{1.548}\\
\cmidrule(lr){1-7}
Improvement & 19.4\%  & 31.2\% & 
34.6\% &   35.0\% & 10.9\%  & 32.1\% \\
\midrule
NS-Informer & 2.094 & 2.710& 0.566&  0.289& 0.861&  1.805  \\
INF-Informer &\textbf{1.487}& \textbf{2.096}&\textbf{0.400}& \textbf{0.249}& \textbf{0.768}&  \textbf{1.739}\\
\cmidrule(lr){1-7}
Improvement& 32.2\%  & 22.7\% &  29.3\%  & 13.8\% &   3.4\% &  3.7\%          \\
\midrule
NS-Autoformer & 1.553&  2.117& 0.580&  0.375& 0.844& 1.923   \\
INF-Autoformer&\textbf{1.427}&  \textbf{2.068}& \textbf{0.421}&  \textbf{0.261}& \textbf{0.794}&\textbf{1.535}\\
\cmidrule(lr){1-7}
Improvement& 8.1\%  & 2.3\%  & 27.7\%   & 
  30.4\% & 5.9\%    &  20.2\%           \\
\bottomrule
\end{tabular}
\end{threeparttable}
}
\vspace{-1mm}
\end{table}

%% file: tables/main_ablation_cr.tex
\begin{table}[!t]
\centering
\small

\caption{Ablation studies taking N-BEATS as the backbone.
} \label{table:appendix_ablation_nbeats}
\vspace{-3mm}
\resizebox{1.0\linewidth}{!}{
\begin{tabular}{c|c|cccccccccc}
\toprule
\multicolumn{2}{c}{Method}& 
\multicolumn{2}{c}{ RealNVP }&\multicolumn{2}{c}{ RealNVP-c}& 
\multicolumn{2}{c}{ IN-Flow-J}&\multicolumn{2}{c}{ IN-Flow-T}& 
\multicolumn{2}{c}{IN-Flow}\\ 
\cmidrule(r){1-2} \cmidrule(r){3-4} \cmidrule(r){5-6} \cmidrule(r){7-8} \cmidrule(r){9-10} \cmidrule(r){11-12} 
\multicolumn{2}{c}{Metric}&
MSE& MAE&  MSE& MAE&
MSE& MAE&  MSE& MAE&
MSE& MAE\\
\midrule \midrule
\multirow{4}{*}{ \rotatebox{90}{Weather}}
&48&1.039&4.973&0.533&2.665&0.495&1.840&0.460&1.731&\textbf{0.432}&\textbf{1.698}\\
&96&1.033&5.354&0.225&1.698&0.234&1.365&0.251&1.376&\textbf{0.223}&\textbf{1.328}\\
&168&0.771&4.724&0.183&1.494&0.186&1.303&0.191&1.243&\textbf{0.182}&\textbf{1.217}\\
&336&0.604&4.647&0.190&1.579&0.183&1.416&0.193&1.382&\textbf{0.179}&\textbf{1.311}\\
\midrule
\multirow{4}{*}{ \rotatebox{90}{NordPool}}
&48&3.137&1.147&1.166&0.643&{1.167}&{0.643}&1.188&0.639&\textbf{1.150}&\textbf{0.632}\\
&96&2.541&1.029&1.602&0.758&1.620&0.759&1.597&0.757&\textbf{1.596}&\textbf{0.756}\\
&168&2.008&0.910&2.006&0.870&1.980&0.845&1.972&0.849&\textbf{1.918}&\textbf{0.836}\\
&336&2.100&0.921&2.283&0.927&{2.152}&{0.898}&2.082&0.880&\textbf{2.098}&\textbf{0.884}\\
\bottomrule
\end{tabular}
}
\vspace{-0mm}
\end{table}

%% file: 6_conclusion.tex
\vfill\eject
\section{Conclusion Remarks} \label{sec:experiments}
In this paper, we propose a principled approach to address the time series distribution shift and contribute to non-stationary time series forecasting. Our core contribution lies in three aspects: a decoupled formulation to separate the removing-shift procedure from the forecasting and regard it as a special distribution transformation, a bi-level optimization problem to formalize the decoupled formulation and enable the joint learning of transformation and forecasting, and a novel invertible network, IN-Flow for time series distribution transformation. 
Moreover, distribution shift is a well-known and crucial topic but is still rarely studied in time series forecasting.
We hope this principled approach, including the decoupled formulation, bi-level optimization, and IN-Flow can facilitate more research on distribution shift in time series and non-stationary  forecasting.

%% file: Appendix.tex
\appendix

\section{More Baseline Details} \label{sec:appendix_baseline}
As aforementioned, our decoupled formulation with {IN-Flow} is model-agnostic such that it can be integrated with any deep time series forecasting models. To verify the effectiveness, we  consider several state-of-the-art models as backbones. Mainly we reproduce the model by following their open source GitHub links.
The detailed implementation details are as follows:
\begin{itemize}
\item
\textbf{iTransformer}~\cite{itransformer_2024} inverts the structure of Transformer without modifying any existing modules by encoding
individual series into variate tokens. These tokens are utilized by the attention mechanism to capture multivariate correlations and FFNs are adopted for each variate token to learn nonlinear representations. 
The official implementation is available at Github\footnote{\url{https://github.com/thuml/iTransformer}}. We follow the default parameters in the original paper in our experiments.

\item \textbf{PatchTST}~\cite{patchtst_2023} divides time series data into subseries-level patches to extract local semantic information and adopts a channel-independence strategy where each channel shares the same embedding and Transformer weights across all the series. The official implementation is available at Github\footnote{\url{https://github.com/yuqinie98/PatchTST}}. We follow the default parameters in the original paper.

\item  \textbf{Autoformer.}~\cite{wu2021autoformer} considers the auto-correlation series and uses auto-correlation to replace the self-attention. We use the official open-source code of Autoformer\footnote{\url{https://github.com/thuml/Autoformer}} \cite{wu2021autoformer}. For hyper-parameter, we follow the settings of the original paper including number of heads, moving average of Auto-correlation layer, dimension of Auto-Correlation layer, dimension of feed-forward layer, etc.
    In addition, we only adopt the positional embeddings and remove time embeddings.
    \item \textbf{N-BEATS.} We take the notable reproduced codes of N-BEATS\footnote{\url{https://github.com/ElementAI/N-BEATS}}. Note that N-BEATS is an univariate time series forecasting model. To align with the input and output of multivariate settings (like Autoformer), we transform the multivariate lookback windows from dimension $B\times L\times D$ to $(B\times L)\times D$, where $B$ is batch size, $L$ is the lookback length and $D$ is number of series. The horizon windows adopt the same strategy. In the mean time, for multivatiate forecasting on different datasets, we set the batch size $B$ equal to $[1024/D]$ to avoid very large batch size that influences training.
    
\end{itemize}
To overcome the distribution shift in time series forecasting, we take state-of-the-art model-agnostic normalization techniques, RevIN \cite{kim2022RevIN}, Dish-TS~\cite{fan2023dish} and SAN~\cite{liu2023adaptive} as the main baselines. We adopt the official implementation\footnote{\url{https://github.com/ts-kim/RevIN}} of RevIN,  the official implementation\footnote{\url{https://github.com/weifantt/Dish-TS}} of Dish-TS, the official implementation\footnote{\url{https://github.com/icantnamemyself/SAN}} of SAN.  We run our models in three different seeds and report the average performances.

We also consider another recent method that's designed for the non-stationarity of transformer-based forecasting models, namely Nonstationary Transformers \cite{liu2022NonstatTranformers} for comparison. To couple with Nonstationary Transformers, we take three model as backbones, including vanilla Transformer \cite{vaswani2017attention}, Informer \cite{zhou2021informer} and Autoformer \cite{wu2021autoformer}. We adopt the official implementation\footnote{\url{https://github.com/thuml/Nonstationary_Transformers/}} of Nonstationary Transformers to set the parameters of nonstationary layers, including nonstationary hidden layers as 2 and hidden size as 128. Specifically,
\begin{itemize}
    \item \textbf{Nonstationary Transformer}. Based on the implementation of vanilla Transformer \cite{vaswani2017attention}, Nonstationary Transformer applies series stationarization layer and revise the raw self attention into the de-stationary self attention \cite{liu2022NonstatTranformers}. Other hyperparameters are set the same as the vanilla Transformer.
    \item \textbf{Nonstationary Informer}. Based on the original implementation\footnote{\url{https://github.com/zhouhaoyi/Informer2020}} of Informer \cite{zhou2021informer}, Nonstationary Informer also adds series stationarization layer and changes the PropSparse attention \cite{zhou2021informer} into the de-stationary PropSparse attention version \cite{liu2022NonstatTranformers}. Other hyperparameters are the same as the Informer.
    \item \textbf{Nonstationary Autoformer}. Since Autoformer replaces classic self attention layers with auto-correlation layers \cite{wu2021autoformer}, Nonstationary Autoformer also adds series stationarization layers and adopts the revised de-stationary auto-correlation layers \cite{liu2022NonstatTranformers} for learning. And other hyperparameters are the same as the Autoformer.
\end{itemize}

Notably, when we integrate the backbones with our IN-Flow, we set all the other hyperparameters the same as the backbones under the same experimental settings (including the random seeds) in order to conduct a fair comparison.

\begin{figure*}[!t]
\centering
\includegraphics[width=0.78\linewidth]{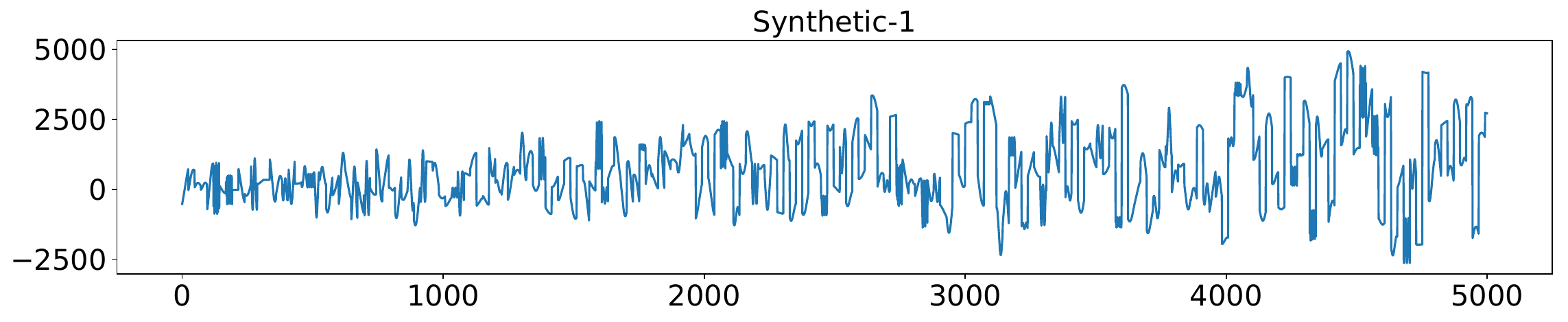} 
\vspace{-2mm}
\caption{An example series of Synthetic-1 dataset.} 
\label{fig:appendix_example_synthetic-1}
\end{figure*}


\begin{figure*}[htbp]
\centering
\subfigure[RealNVP]{
\includegraphics[width=0.17\linewidth]{figs/real.pdf}
\label{fig:appendix_subfig_realnvp}
}
\hspace{+2mm}
\subfigure[RealNVP-c]{
\includegraphics[width=0.17\linewidth]{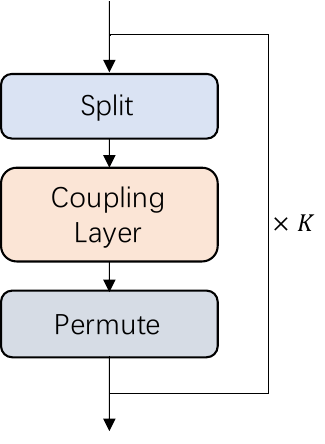}
\label{fig:appendix_subfig_realnvp-c}
}
\hspace{+2mm}
\subfigure[IN-Flow-T (post-norm)]{
\includegraphics[width=0.17\linewidth]{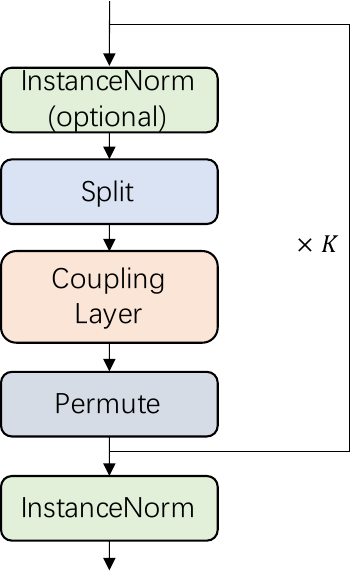}
\label{fig:appendix_subfig_Inflow-T}
}
\hspace{+2mm}
\subfigure[IN-Flow]{
\includegraphics[width=0.2\linewidth]{figs/inflow.pdf}
\label{fig:appendix_subfig_Inflow}
}

\vspace{-2mm}
\caption{Four kinds of variants in ablation studies, where $K$ stands for the number of blocks in flows; ``split'' and ``permute'' stand for two basic operations for coupling layers to assure reversibility \cite{dinh2014nice,dinh2016density}. }
\end{figure*}

\section{More Details on Synthetic Data} \label{sec:appendix_more_syn}

We make two synthetic datasets, namely Synthetic-1 and Synthetic-2, 
We mainly follow the simple rules mentioned in Section \ref{sec:exp_synthetic}, and accordingly evaluate the performance on the synthetic datasets.



Now we illustrate the specific details of the choice of $\mathcal{P}_u$ and $\tau$ for the three datasets.
Specifically, in the $u$-th segment, $s_t =  A_u \operatorname{cos}(2\pi\frac{1}{T_u} t+B_u) + C_u$ where the distribution $\mathcal{P}_u$ of $u$-th segment is controlled by
the amplitude parameters $A_u$, the period parameters $T_u$, the phase parameters $B_u$ and the level parameters $C_u$.
In order to make time series always shifted, we randomly sample the values of $A_u, T_u,B_u,C_u$ to make every segment shifted constantly. We let the amplitude $A_u$ randomly sampled from $(-1000, 1000)$, the period $T_u$ randomly sampled from $(0,100)$, the phase $B_u$ randomly sampled from $(0,100)$ and the level $C_u$ randomly sampled from $(-\lceil t/100 \rceil*50, -\lceil t/100 \rceil*100)$, where $t$ is the timestamp. Then, we can control the distribution changing frequency through adjusting $\tau$. We set $T=10,000$ to synthetic 10,000 points for each series, in order to make enough data for training and evaluation. We take the first 6000 points for training, and 2000 points for validation and another 2000 points for test.
We make Synthetic-1 dataset by setting $\tau=24$ and we make the Synthetic-2 dataset by setting $\tau=12$ in order to model different distribution shift situations. We let each dataset include 5 distinct series, each of which is generated by setting different seeds. Figure \ref{fig:appendix_example_synthetic-1}  demonstrates an example series from the Synthetic-1 dataset respectively, where we show the first five thousand points for visualization.

\section{More Details of Evaluation on Real-world Data} \label{sec:appendix_more_real_world}

\subsection{More Implementation Details}
\label{sec:appendix_more_real_implementation_details}

For a stable evaluation, we rerun all the models under four different seeds and report the average MSE/MAE on the testset of each dataset.
In the main experiments of time series forecasting, we let the lookback window and the horizon window have the same length, where we gradually prolong the length as $\{ 48, 96, 168,  336 \}$ to accommodate short-term/long-term forecasting settings. 
We train all the models using L2 loss and Adam \cite{kingma2014adam} optimizer with learning rate of [1e-4, 1e-3]. The batchsize is set as 1024 for N-BEATS and 128 for others. 
For every model, we use early stopping with the patience as five steps.
For IN-Flow, we traverse the hyperparameters in the validation set: with the number of blocks in coupling layers as $\{ 2, 8, 16 \}$, hidden size as 128. The learning rate for the transformation module is set as 1e-4. All the experiments are implemented with PyTorch \cite{paszke2019pytorch} on single NVIDIA A100 40GB GPU.

\vspace{-1mm}
\subsection{More Evaluation Metric Details}
\label{sec:appendix_more_real_evaluation_details}

To directly reflects the shift in time series, all experiments and evaluations are conducted on original data without any data normalization or data scaling, which is different from experimental settings of \cite{zhou2021informer,wu2021autoformer} which use z-score normalization to process data before training and evaluation.


In evaluation, we evaluate the time series forecasting performances on the mean squared error (MSE) and mean absolute error (MAE). Note that our experiments and evaluations are on original data; thus the reported metrics on real-world data are scaled for readability.
Specifically, for MSE, the report metrics are scaled by $1e^{-1}$ on \textit{ETTh1} dataset and \textit{ETTm2} dataset, 
by $1e^{-4}$ on \textit{Weather} and \textit{Electricity} dataset, by $1e^{-6}$  on \textit{CAISO} dataset and by $1e^{-6}$ on \textit{NordPool} dataset. For MAE, the reported metrics are not scaled for ETT series datasets. In contrast, the reported MAE is scaled by $1e^{-1}$ on \textit{Weather} dataset, by $1e^{-3}$ on \textit{CAISO} dataset and by $1e^{-2}$ on \textit{NordPool} dataset.